\documentclass[conference]{IEEEtran}
\IEEEoverridecommandlockouts

\usepackage[T1]{fontenc}
\usepackage{cite}
\usepackage{amsmath,amssymb,amsfonts}
\usepackage{graphicx}
\usepackage{booktabs}
\usepackage{multirow}
\usepackage{url}
\usepackage[hidelinks]{hyperref}
\usepackage{xcolor}
\usepackage{microtype}
\usepackage{placeins}

\begin{document}
\bstctlcite{IEEEexample:BSTcontrol}

\title{Boundary Fragility in Chunked Robot Manipulation and Post-Training}

\author{
\IEEEauthorblockN{Qiancheng Zhou}
\IEEEauthorblockA{\textit{School of Future Technology, Shanghai University}, Shanghai, China \\
\texttt{zhou\_3721@shu.edu.cn}}
}

\maketitle

\begin{abstract}
Foundation models for robot manipulation bridge slow inference and fast control by executing a chunk of future actions open-loop per query, yet frequently fail during fine contact, unseen resistance, and retries. Existing remedies assume open-loop errors accumulate uniformly over time, globally shortening horizons or averaging overlapping predictions. We show that this \emph{boundary fragility} arises not from uniform drift across smooth motion regimes, which remain locally contractive, but specifically across \textbf{physical contact boundaries} and \textbf{decision branching boundaries}, where targeted interventions suffice without shortening smooth interiors. During execution, lag induces slip on curved rims while initial finger-pad compression masks unseen joint friction; bounding delayed gain before touch and probing post-compression displacement with closure withdrawal lifts asynchronous completion from $41.4\%$ to $90.6\%$ and cuts peak backstop force from $452.9\,\mathrm{N}$ to $183.6\,\mathrm{N}$. Once contact exposes free versus jammed motion, cross-boundary regression collapses onto the majority route; truncating supervision on this feedback boundary with contact replanning raises flow-matching jam recovery from $35.0\%$ to $88.3\%$ while leaving non-branching tasks in \textsc{LIBERO-SelfTest} unchanged. Yet even when imitation preserves multiple routes, reinforcement post-training leaves retries repeating blocked motions by collapsing only the opening branch choice while keeping within-branch execution intact; running the pre-RL policy solely across the opening steps restores 7B VLA $3\mathrm{D}$ approach coverage from $0.227$ to $0.342$ with full accuracy ($0.675$). Code and \textsc{LIBERO-SelfTest} are available at \url{https://github.com/ershiyidian/libero-selftest} and \url{https://huggingface.co/datasets/esyd/libero-selftest}.
\end{abstract}

\section{Introduction}
\label{sec:intro}

General-purpose robot manipulation increasingly builds on foundation models, spanning end-to-end vision-language-action (VLA) policies~\cite{openvla2024,oft2025,smolvla2025,pi05_2025}, native vision-language robot policies~\cite{astranomy2026,astra_robodojo2026}, and hierarchical agentic stacks in which vision-language planners steer low-level VLA primitives~\cite{harnessvla2026,pyrualean2026,interactvla2026}. Because foundation-model inference takes tens of milliseconds to several seconds whereas motor control runs at dozens of hertz, practical policies emit a short sequence of future motor increments per query---an \emph{action chunk}~\cite{act2023}---and execute it open-loop until the next update. Despite strong semantic understanding and high nominal benchmark scores, broad evaluations on RoboDojo~\cite{robodojo2026}, LIBERO-Plus~\cite{liberoplus2026}, and LIBERO-RECOVER~\cite{libero_recovery} expose a persistent deployment gap: standalone VLAs average under $9\%$ task success on multi-stage RoboDojo manipulation, and even agent-guided stacks frequently break down during fine contact alignment, unseen mechanical resistance, and post-failure retries~\cite{astra_robodojo2026,interactvla2026}.

\begin{figure*}[t]
\centering
\includegraphics[width=0.96\textwidth]{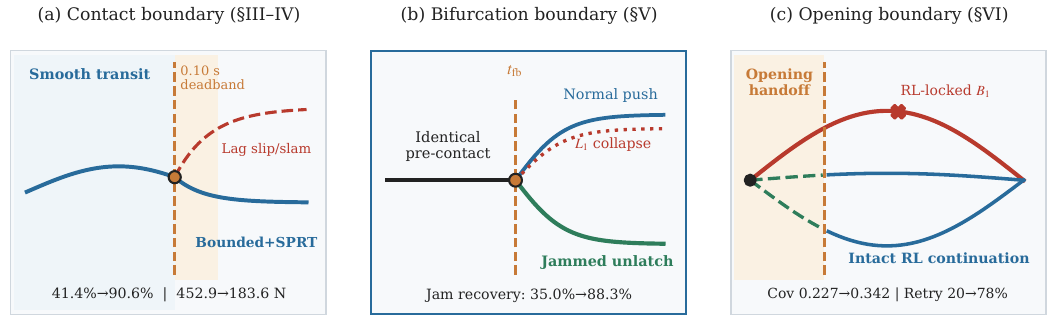}
\caption{\textbf{Boundary fragility across chunked manipulation and post-training.} (a)~In physical execution (\S\ref{sec:approach}--\S\ref{sec:probing}), free-space transit is naturally contractive, whereas lag overshoot past curved-rim self-locking limits and the $0.10\,\mathrm{s}$ finger-pad compliance deadband concentrate failure along contact transitions. (b)~In imitation learning (\S\ref{sec:supervision}), identical pre-contact states branch past feedback boundary $t_{\mathrm{fb}}$ into normal pushing versus jammed unlatching; unmasked $L_1$ chunks collapse onto the majority branch, whereas causal boundary truncation paired with flow matching and contact replanning recovers jammed trials. (c)~In RL post-training (\S\ref{sec:rl_narrowing}), outcome rewards lock only the opening branch boundary ($B_1$) while leaving continuation inside suppressed corridors intact, enabling zero-retraining opening handoff to unblock retries.}
\label{fig:teaser}
\vspace{-4pt}
\end{figure*}

Existing remedies treat open-loop vulnerability as a smooth degradation across time or internal model confidence. Execution-time controllers average overlapping chunks~\cite{act2023}, inpaint or condition on committed asynchronous prefixes~\cite{rtc2025,training_rtc2025}, route queries across a known monotone phase transition~\cite{tract2026}, or shorten horizons before contact using speed valleys~\cite{pace2026}, denoising variance~\cite{dvac2026}, self-attention plateaus~\cite{autohorizon2026}, or imagined world-model rollouts~\cite{ca3c2026}. However, multi-step execution is not uniformly fragile along a trajectory. Within free-space transit or a single smooth physical regime, closed-loop tracking and learned continuation skills are naturally contractive; breakdowns concentrate sharply when a committed sequence crosses a \textbf{physical contact boundary} or a \textbf{decision branching boundary} (Fig.~\ref{fig:teaser}). Treating smooth interiors and sharp boundaries alike sacrifices throughput where long open-loop commitments are safe, yet still fails where contact mechanics jump or post-feedback trajectories diverge.

Following a manipulation trajectory chronologically---from pre-contact approach, through initial touch and post-contact continuation, to multi-attempt retry---shows how these boundaries form a single coupled chain rather than isolated failure modes across discrete-skill (QueST~\cite{quest2024}), continuous regression (OpenVLA-OFT~\cite{oft2025}), and flow-matching (SmolVLA~\cite{smolvla2025}) policies in $500\,\mathrm{Hz}$ MuJoCo physics across LIBERO~\cite{libero2023}. During approach, inference lag cycles in a periodic sawtooth pattern within each chunk, remaining harmless in free space until crossing a periodic stability limit where terminal overshoot pushes parallel-jaw grasps outside curved-rim friction cones into upward slip (\S\ref{sec:approach}). Once touch is established on an articulated handle, pre-contact visual images and internal model attention or denoising statistics carry strictly zero Fisher information about hidden joint resistance, while elastic finger-pad compression creates a brief post-contact deadband before low- and high-friction displacements separate and rigid closure risks severe backstop slam (\S\ref{sec:probing}). Crucially, exposing whether a joint moves freely or jams at contact creates an immediate information boundary for policy supervision: identical pre-contact observations branch after touch into forward pushing versus unlatching withdrawal. Supervising an unmasked chunk across that feedback instant forces deterministic regression onto the majority branch, while unconditional asynchronous prefix skipping discards unlatch prefixes whenever a jammed joint immobilizes the gripper during inference---explaining why chunk truncation helps if and only if execution crosses an informative post-contact bifurcation (\S\ref{sec:supervision}).

Yet preserving local recovery branches during imitation learning still does not guarantee resilient deployment across multi-attempt retries. Modern post-training pipelines routinely apply online outcome reinforcement learning (RIPT-VLA~\cite{ript2025}, SimpleVLA-RL~\cite{simplevlarl2026}) to sharpen both visuomotor VLAs and symbolic subgoal planners, raising single-rollout precision at the cost of global behavioral contraction and trajectory homogenization~\cite{drive2026,extoken2026}: when an obstacle or jammed mechanism blocks the dominant route, subsequent retries repeat nearly identical failed motions~\cite{interactvla2026}. By analyzing two-stage KL-regularized gradient flow across 7B OpenVLA-OFT and exhaustively enumerable symbolic planning~\cite{pan2025countdown}, we find that outcome optimization does not degrade downstream execution inside alternative branches. Instead, terminal rewards concentrate probability mass strictly at the \textbf{opening branch boundary} while shared continuation parameters keep improving across unvisited routes (\S\ref{sec:rl_narrowing}). Because within-branch competence remains intact, policies need not be retrained from scratch with auxiliary trajectory-diversity rewards~\cite{drive2026,extoken2026}; simply handing the opening segment back to the pre-RL policy recovers multi-corridor breadth and unblocks closed-loop retries without sacrificing RL accuracy.

In summary, our contributions span three lifecycle boundaries:
\begin{enumerate}
    \item \textbf{Physical Contact and Probing Boundaries in Execution (\S\ref{sec:approach}--\S\ref{sec:probing}):} Without world models or architectural changes, we derive periodic Floquet stability limits and curved-rim self-locking bounds that pull delayed asynchronous grasps inside the self-locking disk (lifting completion from $41.4\%$ to $90.6\%$), and pair post-deadband Wald sequential probing with closure target withdrawal across $800$ articulated-joint rollouts to resolve unseen friction in $3.4$ steps while cutting peak backstop force from $452.9\,\mathrm{N}$ to $183.6\,\mathrm{N}$---surpassing pre-contact adaptive-horizon baselines that remain below $58\%$ under high resistance.
    \item \textbf{Causal Chunk Imitation Across Post-Contact Bifurcations (\S\ref{sec:supervision}):} We prove why post-contact branching collapses unmasked $L_1$ chunks onto the majority mode and introduce \textsc{LIBERO-SelfTest} across all $130$ LIBERO tasks ($1{,}040$ replay-verified pass and recovery episodes). Pairing causal boundary truncation with conditional flow matching and contact-triggered replanning lifts jammed recovery from $35.0\%$ to $88.3\%$---and restores delayed latch recovery on pretrained SmolVLA-450M via displacement-gated prefix alignment---while confirming no gain on non-branching single-physics tasks.
    \item \textbf{Opening-Boundary Localization and Zero-Retraining Retry Recovery in RL Post-Training (\S\ref{sec:rl_narrowing}):} Auditing two-stage gradient dynamics across 7B OpenVLA-OFT and enumerable symbolic reasoning establishes that outcome RL narrows initial branch selection rather than continuation execution. Without diversity-regularized RL retraining~\cite{drive2026}, running the pre-RL policy solely across the opening steps restores 7B VLA $3\mathrm{D}$ corridor coverage from $0.227$ to $0.342$ at full single-try accuracy ($0.675$) and lifts $200\,\mathrm{ms}$ closed-loop jammed-subgoal retry completion from $20.0\%$ to $78.0\%$.
\end{enumerate}

\section{Related Work}
\label{sec:related_work}

\subsection{Multi-Step Action Chunking, Latency, and Adaptive Horizons}
Generalist manipulation policies predict multi-step action chunks via discrete skill tokens~\cite{quest2024,fast2025}, continuous regression~\cite{act2023,oft2025}, or conditional flow matching~\cite{pi02024,pi05_2025,smolvla2025}, and are increasingly paired with or complemented by high-latency vision-language agents acting either natively~\cite{astranomy2026,astra_robodojo2026} or over low-level VLA primitives~\cite{harnessvla2026,pyrualean2026,interactvla2026}. Existing execution remedies address open-loop drift along two lines. Time-uniform controllers smooth overlapping predictions (ACT~\cite{act2023}), inpaint or condition on committed asynchronous prefixes (RTC~\cite{rtc2025}, Training-Time RTC~\cite{training_rtc2025}, EAPN~\cite{eapn2026}), or compress flow-sampling steps (FASTER~\cite{faster2026}), treating all steps along a trajectory alike. Adaptive-horizon controllers instead shorten commitment before acting using kinematic speed valleys (PACE~\cite{pace2026}), final-step denoising variance (DVAC~\cite{dvac2026}), action self-attention plateaus (AutoHorizon~\cite{autohorizon2026}), or disagreement among imagined trajectories in an action-conditioned world model ($\text{CA}^3\text{C}$~\cite{ca3c2026}). Without learned world models or architectural changes, we derive analytical periodic stability limits and curved-contact self-locking bounds that pinpoint when delayed chunks cross from contractive free-space motion into grasp slip, and identify why unconditional prefix skipping discards recovery motion under physical stall.

\subsection{Active Contact Probing and Cross-Bifurcation Recovery Learning}
Recent contact-rich policies condition their backbones on continuous high-dimensional wrist force-torque or tactile streams~\cite{forcevla2025,tacvla2026,forceflow2026}, whereas interactive perception infers latent mechanics from active kinematic response~\cite{interactive_perception2017} without tactile or force encoders. Scripted probing routines for vision-language agents (PhysCaP~\cite{physcap2026}) use fixed lift or squeeze primitives to estimate mass and stiffness, overlooking two boundaries of continuous joint actuation: compliant finger pads create an initial zero-information compression deadband before static breakaway, and holding forward targets past mechanical closure causes severe backstop impacts. Meanwhile, perturbation and recovery benchmarks (RoboDojo~\cite{robodojo2026}, LIBERO-Plus~\cite{liberoplus2026}, LIBERO-RECOVER~\cite{libero_recovery}) show that high nominal task success masks poor recovery from interaction-induced contact deviations. Closest to our training analysis, TRACT~\cite{tract2026} resolves phase mismatch when an action chunk crosses a procedural boundary, but assumes a known monotone transition to a unique successor phase. When unseen joint resistance instead creates mutually exclusive post-contact branches (normal continuation versus jammed withdrawal), pre-contact visual observations and internal model confidence carry zero Fisher information. We resolve latent resistance online via post-deadband sequential testing~\cite{wald1945sequential} and prove that imitation requires truncating supervision strictly at the contact-feedback boundary alongside multimodal flow matching and event-triggered replanning.

\subsection{VLA Reinforcement Post-Training and Multi-Route Retry}
Beyond demonstration cloning, online reinforcement fine-tuning (RIPT-VLA~\cite{ript2025}, SimpleVLA-RL~\cite{simplevlarl2026}, RLinf-VLA~\cite{rlinfvla2026}) improves single-rollout VLA and planner precision from task outcomes (mirroring GRPO in symbolic reasoning~\cite{shao2024deepseekmath,pan2025countdown}), yet induces global behavioral contraction and rollout homogenization~\cite{drive2026,extoken2026} that leaves multi-attempt retries repeating identical blocked motions~\cite{interactvla2026}. Recent methods counter this narrowing during training by routing rollouts through discrete exploration tokens (ExToken~\cite{extoken2026}) or shaping rewards with temporally aligned trajectory diversity among successful rollouts (DRIVE~\cite{drive2026}), both requiring modified reinforcement training loops. Rather than retraining with auxiliary exploration objectives, we show that outcome reinforcement learning narrows selection specifically at the opening branch boundary while leaving downstream execution within suppressed branches intact, enabling zero-retraining opening handoff to recover spatial corridor coverage and closed-loop retry success.

\section{Approach Latency and Curved-Contact Boundaries}
\label{sec:approach}

\begin{figure*}[t]
\centering
\includegraphics[width=0.92\textwidth]{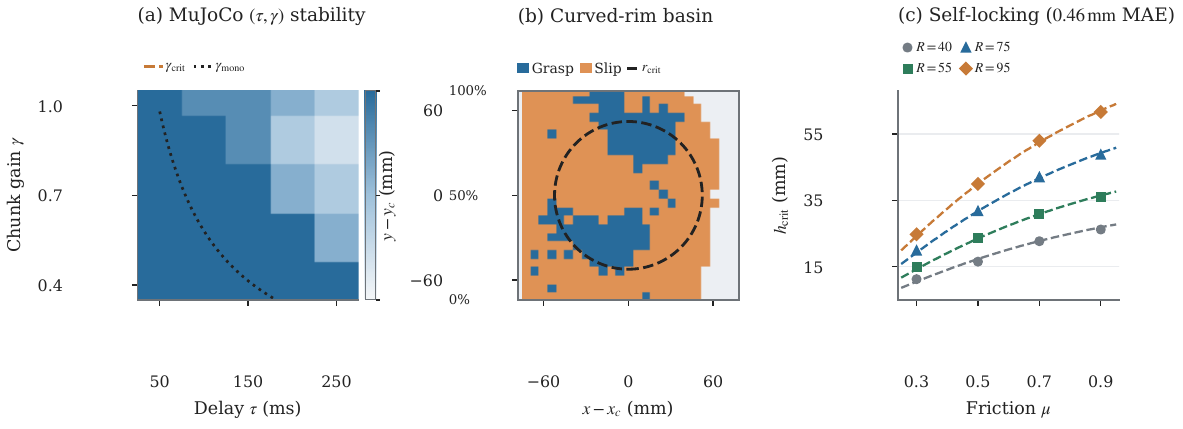}
\caption{Approach delay stability and curved-rim self-locking boundaries. (a)~MuJoCo $(\tau,\gamma)$ completion phase map overlaid with analytical monotonic-damping $\gamma_{\mathrm{mono}}(d)$ and stability $\gamma_{\mathrm{crit}}(d)$ curves. (b)~Bowl-centered $25\times 25$ grasp basin ($x-x_c, y-y_c \in [-72,72]\,\mathrm{mm}$) with the analytical Coulomb self-locking circle $r_{\mathrm{crit}} = 52.3\,\mathrm{mm}$. (c)~Controlled $500\,\mathrm{Hz}$ MuJoCo verification across four smooth curvatures and four friction coefficients ($576$ trials, $0.46\,\mathrm{mm}$ MAE against Eq.~\eqref{eq:friction-cone}).}
\label{fig:approach-contact}
\vspace{-4pt}
\end{figure*}

\subsection{Periodic Delay Stability and Curved Self-Locking}
\label{sec:stability}

Under inference and queueing delay, commands executed at the current control step originate from an earlier observation. This lag creates an opposing requirement between free-space motion and terminal contact alignment. During steady transit, the arm and target move together so unlagged incremental commands cancel velocity drift, whereas anchoring to a historical pose pulls curved paths backward toward obstacles. Near terminal docking, however, the target settles while the arm is still moving; uncalibrated increments then accumulate forward overshoot, whereas anchoring to the historical enqueue pose or skipping expired prefix steps suppresses terminal drift.

Executing a multi-step action chunk open-loop compounds this lag periodically: observation age grows step by step within each chunk and resets at the next policy query, forming a periodic sawtooth delay. Lifting the resulting tracking error over one chunk stride defines a periodic linear system governed by its Floquet contraction factor~\cite{floquet1883equations}. To transition smoothly from undamped transit to damped contact alignment without manual phase switches, we bound the open-loop chunk gain within the periodic Floquet stability limit and blend current and enqueue pose references by the ratio of observed arm speed to commanded chunk speed:
\begin{equation}
\begin{aligned}
\bar{\rho}(d_0, c, \gamma) &= \rho\!\left(\prod_{k=0}^{c-1} \mathbf{A}_{d_0+k}(\gamma)\right)^{1/c} < 1, \\
g_n^{\mathrm{cl}} &= (1 - w_n)\bigl(x_n + \gamma_{\mathrm{stab}}\Delta g_{n-d}\bigr) \\
&\quad + w_n\bigl(x_{n-d} + \Delta g_{n-d}\bigr),
\end{aligned}
\label{eq:floquet-blend}
\end{equation}
where $d_0$ is the base inference lag, $c$ is the executed chunk stride, $\mathbf{A}_{d_0+k}(\gamma)$ is the one-step delayed error operator under gain $\gamma$, $\gamma_{\mathrm{stab}}$ enforces $\bar{\rho} < 1$, and $w_n \in [0, 1]$ weights the historical enqueue pose $x_{n-d}$ against the current pose $x_n$ as commanded chunk speed vanishes relative to observed arm speed (Proposition~1 in Appendix~\ref{app:prop-delay}).

Approach overshoot rarely causes failure in free space; instead, execution breaks specifically when jaw closure meets curved object geometry. When radial approach error shifts parallel-jaw clamping force $f_c$ onto a flared rim of local wall angle $\varphi(r)$ and friction coefficient $\mu$, resolving normal and tangential contact forces yields the net vertical ejection force on an object of mass $m$:
\begin{equation}
F_z(r) = 2 f_c \bigl(\sin\varphi(r) - \mu\cos\varphi(r)\bigr) - m g, \quad \tan\varphi(r_{\mathrm{crit}}) = \mu.
\label{eq:friction-cone}
\end{equation}
Equation~\eqref{eq:friction-cone} isolates a sharp geometric contact boundary: inside the self-locking radius $r_{\mathrm{crit}}$, friction sustains lift, whereas landing outside $r_{\mathrm{crit}}$ wedges the object upward out of the gripper or twists it into slip under single-finger wrist-yaw misalignment (yaw-wrench mechanics in Appendix~\ref{app:mechanics}; multi-curvature and concave-handle audits in Appendix~\ref{app:ext-mechanics}).

\subsection{Validation Across VLA Action Families}
\label{sec:exp-delay}

\begin{table*}[t]
\centering
\caption{Closed-Loop Task Completion (\%) Across Three VLA Action Paradigms Under Observation Delay.}
\label{tab:delay}
\scriptsize
\setlength{\tabcolsep}{4.6pt}
\begin{tabular}{lll c ccccc ccccc}
\toprule
& & & \multicolumn{1}{c}{$0\,\mathrm{ms}$} & \multicolumn{5}{c}{$100\,\mathrm{ms}$ ($d = 2$, $\bar\rho = 0.9588$)} & \multicolumn{5}{c}{$200\,\mathrm{ms}$ ($d = 4$, $\bar\rho = 1.0012$)} \\
\cmidrule(lr){4-4} \cmidrule(lr){5-9} \cmidrule(lr){10-14}
Model & Suite & Train. & Raw & Raw & Iss. & ACT-TE & Pref.-Al. & Fl.-Blend & Raw & Iss. & ACT-TE & Pref.-Al. & Fl.-Blend \\
\midrule
\multicolumn{14}{l}{\emph{Discrete-skill VLA: QueST ($500$ official episodes/row; $H = 8$ stride)}} \\
QueST & Goal & RIPT & 86.2 & 47.0 & 86.6 & 60.0 & \textbf{94.0} & 58.0 & 19.0 & 80.4 & 26.0 & 86.0 & \textbf{88.0} \\
QueST & Goal & SFT & 79.2 & 51.0 & 84.2 & 74.0 & \textbf{90.0} & 66.0 & 31.0 & 85.6 & 42.0 & \textbf{96.0} & 94.0 \\
QueST & Spatial & RIPT & 92.0 & 32.0 & \textbf{95.6} & 28.0 & 90.0 & 32.0 & 2.2 & 90.8 & 0.0 & 88.0 & \textbf{98.0} \\
QueST & Spatial & SFT & 85.6 & 44.8 & \textbf{92.6} & 52.0 & 90.0 & 54.0 & 16.6 & 93.4 & 22.0 & \textbf{94.0} & 92.0 \\
\cmidrule(lr){1-14}
QueST mean & 2 suites & 4 ckpts & 85.8 & 43.7 & 89.8 & 53.5 & \textbf{91.0} & 52.5 & 17.2 & 87.6 & 22.5 & 91.0 & \textbf{93.0} \\
\midrule
\multicolumn{14}{l}{\emph{Continuous $L_1$ 7.65B VLA: OpenVLA-OFT ($500$ official episodes/row; $H = 8$; $300\,\mathrm{ms}$ perturbed audit in Appendix~\ref{app:ext-async})}} \\
OpenVLA-OFT & Spatial & RIPT & 98.8 & 87.2 & 98.4 & 75.0 & \textbf{100.0} & 85.0 & 43.2 & 95.0 & 35.0 & \textbf{100.0} & 90.0 \\
OpenVLA-OFT & Spatial & SFT & 98.2 & 87.8 & \textbf{98.2} & 85.0 & 95.0 & 95.0 & 44.0 & 96.0 & 50.0 & \textbf{100.0} & \textbf{100.0} \\
\midrule
\multicolumn{14}{l}{\emph{Flow-matching VLA: SmolVLA-450M ($20$-episode diagnostic subset/suite; $50$-step horizon; $2{,}880$-rollout $300\,\mathrm{ms}$ audit in Appendix~\ref{app:ext-async})}} \\
SmolVLA mean & 4 suites & Flow & 45.0 & 36.2 & 40.0 & 38.8 & \textbf{45.0} & 37.5 & 20.0 & 20.0 & 25.0 & \textbf{46.2} & 40.0 \\
\bottomrule
\end{tabular}
\vspace{2pt}

\parbox{\linewidth}{\scriptsize\emph{Note:} Iss.\ ($m=d$ enqueue pose); ACT-TE~\cite{act2023} (decay $0.01$); Pref.-Al.~\cite{rtc2025} ($i=d$); Fl.-Blend (Eq.~\eqref{eq:floquet-blend}, $\gamma_{\mathrm{stab}}=1.00$ at $100\,\mathrm{ms}$, $0.51$ at $200\,\mathrm{ms}$). Bold marks best per row at each delay.}
\end{table*}

We evaluate delayed closed-loop execution in $500\,\mathrm{Hz}$ MuJoCo physics~\cite{todorov2012mujoco} across LIBERO~\cite{libero2023} spanning discrete-skill (QueST~\cite{quest2024}), continuous regression (OpenVLA-OFT~\cite{oft2025}), and flow-matching (SmolVLA~\cite{smolvla2025}) policies (Table~\ref{tab:delay}, Fig.~\ref{fig:approach-contact}a). At $100\,\mathrm{ms}$ delay ($d=2$), unit-gain execution remains inside the periodic stability envelope ($\bar{\rho}=0.9588<1$, so $\gamma_{\mathrm{stab}}=1.00$ is inactive): full enqueue calibration (\textsc{Iss.}, $w_n \equiv 1$, $89.8\%$ on QueST) or prefix skipping (RTC~\cite{rtc2025}, $91.0\%$) removes phase lag cleanly, whereas Eq.~\eqref{eq:floquet-blend}'s continuous speed-ratio weight $w_n \in [0,1]$ retains $(1-w_n)$ uncalibrated current-pose increments whenever QueST's discrete VQ skills decelerate mid-transit, leaving residual phase drift ($52.5\%$ vs.\ $43.7\%$ Raw). Once delay reaches $200\,\mathrm{ms}$ ($\bar{\rho}>1$, $\gamma_{\mathrm{stab}}=0.51$), however, undamped pose calibration destabilizes ($17.2\%$ Raw on QueST, $20.0\%$ on SmolVLA) and chunk averaging (ACT-TE~\cite{act2023}) blurs discrete skills ($22.5\%$), whereas periodic gain damping becomes decisive---rebounding QueST to $93.0\%$ and doubling SmolVLA to $40.0\%$ ($46.2\%$ with prefix skipping; $75.8\%$ combined at $300\,\mathrm{ms}$ and $92.0\%$ on 7B OpenVLA-OFT in Appendix~\ref{app:ext-async})---motivating selective activation of Eq.~\eqref{eq:floquet-blend} only once $\bar{\rho}(d_0,c,1) \ge 1$.

Controlled contact mapping on curved bowls confirms that crossing $r_{\mathrm{crit}}$ is the physical cause of these delayed failures (Fig.~\ref{fig:approach-contact}b--c). Across four surface curvatures and four friction levels ($576$ trials), empirical slip boundaries match Eq.~\eqref{eq:friction-cone} within $0.46\,\mathrm{mm}$ mean absolute error, concentrating grasp success inside the $52.3\,\mathrm{mm}$ self-locking disk by $3.15\times$. Across $256$ asynchronous wall-clock VLA rollouts, bounding periodic chunk gain pulls the median first-contact radius from the outer slip ring back inside $r_{\mathrm{crit}}$ ($70.4\,\mathrm{mm}$ to $51.8\,\mathrm{mm}$) and raises task completion from $41.4\%$ to $90.6\%$. Stabilizing the approach boundary thus lands the gripper reliably on target geometry, yet pre-contact visual alignment still reveals nothing about internal mechanical resistance once contact begins.

\section{Post-Contact Deadband and Sequential Friction Probing}
\label{sec:probing}

\begin{figure*}[t]
\centering
\includegraphics[width=0.92\textwidth]{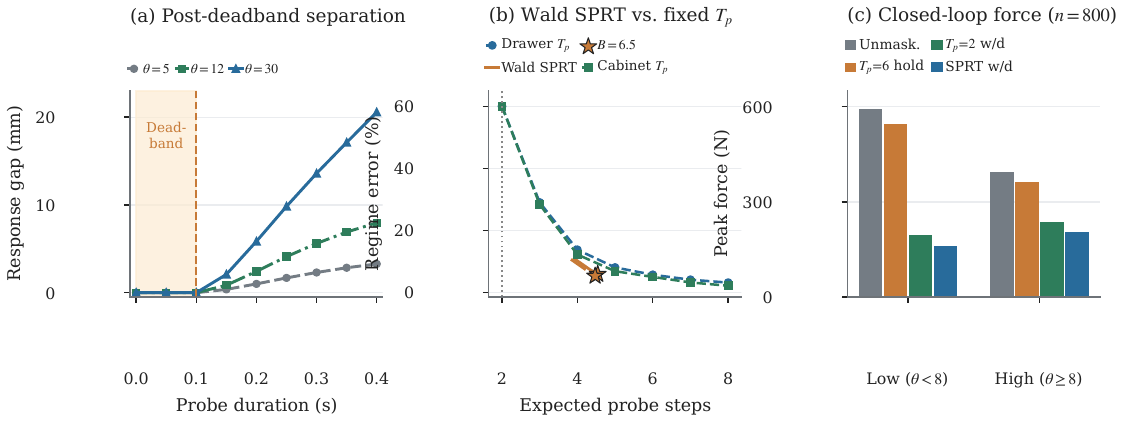}
\caption{Post-deadband displacement separation, online Wald SPRT probing, and closure-triggered target withdrawal. (a)~End-effector response gap versus probe duration showing the $2$-step ($0.10\,\mathrm{s}$) finger-pad compliance deadband followed by monotonic separation across hidden friction levels $\theta$. (b)~Regime classification error versus expected probe steps across ten friction levels on articulated drawer and cabinet fixtures comparing fixed horizons $T_p \in \{2,\dots,8\}$ against online Wald SPRT ($B \in \{2.5, 4.5, 6.5\}$). (c)~Mean peak contact force across $800$ closed-loop MuJoCo episodes.}
\label{fig:probing-deadband}
\vspace{-4pt}
\end{figure*}

\subsection{Pad-Compliance Deadband and Online Sequential Probing}
Once the gripper contacts an articulated drawer or cabinet joint of unseen Coulomb friction, continuous actuation encounters two post-contact mechanical boundaries. First, tapering pushing force before the joint reaches mechanical closure causes high-friction joints to stall, whereas maintaining forward chunk targets past mechanical closure slams the rigid backstop above $450\,\mathrm{N}$. Retracting the command target along the push axis immediately upon detecting closure eliminates backstop impact without sacrificing stroke completion.

Second, pre-contact visual images are identical across hidden friction levels and therefore carry strictly zero Fisher information about internal resistance. Even after touch occurs, compliant finger pads deform elastically before static breakaway begins. During this initial pad-compression deadband ($d_{\mathrm{dead}}$ steps), end-effector displacement is governed by pad stiffness rather than joint friction, keeping Fisher information identically zero (Proposition~2 in Appendix~\ref{app:prop-deadband}). Only past $d_{\mathrm{dead}}$ do low- and high-friction displacement trajectories separate (Fig.~\ref{fig:probing-deadband}a). We therefore accumulate post-deadband displacement evidence online using Wald's Sequential Probability Ratio Test (SPRT)~\cite{wald1945sequential} between calibrated low- and high-friction displacement prototypes $(\mu_j^{\mathrm{low}}, \mu_j^{\mathrm{high}})$ under measurement variance $\sigma_y^2$:
\begin{equation}
\begin{aligned}
\Lambda_m &= \frac{1}{2\sigma_y^2} \sum_{j = d_{\mathrm{dead}}+1}^{m} \Bigl( \bigl\| y_j - \mu_j^{\mathrm{low}} \bigr\|_2^2 - \bigl\| y_j - \mu_j^{\mathrm{high}} \bigr\|_2^2 \Bigr), \\
N_{\mathrm{SPRT}} &= \min\bigl\{ m > d_{\mathrm{dead}} : |\Lambda_m| \ge B \;\text{or}\; m = T_{\max} \bigr\},
\end{aligned}
\label{eq:wald-sprt}
\end{equation}
where $y_j$ is the observed displacement at probe step $j$, $B$ is the decision threshold, and $T_{\max}$ caps the probe horizon. Equation~\eqref{eq:wald-sprt} exits in $3$ to $4$ steps when resistance is clearly low or high while allocating extra steps near the decision boundary.

\subsection{Closed-Loop Probing vs.\ Pre-Contact Adaptive Horizons}

\begin{table}[t]
\centering
\caption{Closed-Loop Joint Probing, Closure Withdrawal, and Pre-Contact Adaptive Horizons ($2{,}720$ MuJoCo Rollouts).}
\label{tab:probing}
\scriptsize
\setlength{\tabcolsep}{2.6pt}
\begin{tabular}{lccccc}
\toprule
\multicolumn{6}{l}{\emph{A. Articulated drawer \& cabinet sweep ($800$ rollouts, $10$ friction levels)}} \\
Controller & Acc.\ (\%) & $T_p$ & Low / High & Peak ($\mathrm{N}$) & Ovld$\downarrow$ \\
\midrule
Unmasked cross-boundary & -- & 0.00 & 100.0 / 90.0 & 494.3 & -- \\
Fixed short ($T_p = 2$) + w/d & 76.2 & 2.00 & 93.8 / 77.5 & 217.0 & -- \\
Fixed long ($T_p = 6$), hold & 85.6 & 6.00 & 100.0 / 90.0 & 452.9 & -- \\
Fixed long ($T_p = 6$) + w/d & 84.4 & 6.00 & 100.0 / 90.0 & 187.4 & -- \\
\textbf{Online Wald SPRT + w/d} & \textbf{86.9} & \textbf{3.39} & \textbf{100.0} / 85.0 & \textbf{183.6} & -- \\
\midrule
\multicolumn{6}{l}{\emph{B. SmolVLA-450M under unseen resistance ($1{,}920$ rollouts, $5$ fixtures)}} \\
Fixed-$16$ baseline & -- & 0.00 & 85.8 / 57.5 & 150.3 & 25.0 \\
PACE~\cite{pace2026} (speed valley) & -- & 0.00 & 85.0 / 53.3 & 153.2 & 23.8 \\
DVAC~\cite{dvac2026} (denoising var.) & -- & 0.00 & 81.7 / 55.8 & 151.8 & 23.3 \\
AutoHorizon~\cite{autohorizon2026} (attn.) & -- & 0.00 & 85.8 / 57.5 & 153.3 & 26.7 \\
Fixed ($T_p = 6$), hold & 83.8 & 6.00 & \textbf{85.8 / 79.2} & 163.4 & 32.5 \\
\textbf{Online Wald SPRT + w/d} & \textbf{84.7} & \textbf{4.59} & 80.0 / 69.2 & \textbf{109.1} & \textbf{5.0} \\
\bottomrule
\end{tabular}
\vspace{2pt}

\parbox{\linewidth}{\scriptsize\emph{Note:} w/d denotes $4\,\mathrm{mm}$ closure withdrawal ($d_{\mathrm{dead}}=2$). Panel~B evaluates $F_{\mathrm{res}} \in [2,78]\,\mathrm{N}$ with $180\,\mathrm{N}$ overload threshold (compound $200$--$300\,\mathrm{ms}$ lag and viscous-spring audits in Appendix~\ref{app:ext-probing}).}
\end{table}

Across $800$ closed-loop rollouts spanning ten friction levels on articulated drawer and cabinet fixtures (Table~\ref{tab:probing}A, Fig.~\ref{fig:probing-deadband}b--c), fixed short probes that terminate inside the $0.10\,\mathrm{s}$ pad-compression deadband incur up to $60.0\%$ offline classification error and drop high-friction completion to $77.5\%$, whereas premature force tapering stalls completely on high-resistance joints. Conversely, holding forward targets without withdrawal produces $452.9\,\mathrm{N}$ to $494.3\,\mathrm{N}$ backstop spikes. Pairing post-deadband sequential probing with closure-triggered target withdrawal resolves friction regimes at $86.9\%$ accuracy in $3.39$ mean steps while cutting peak contact force to $183.6\,\mathrm{N}$.

Evaluating recent training-free adaptive-horizon controllers on SmolVLA-450M across unseen joint resistance further highlights why pre-contact introspection cannot substitute for physical probing (Table~\ref{tab:probing}B). Controllers that shorten chunks from speed valleys (PACE~\cite{pace2026}), denoising variance (DVAC~\cite{dvac2026}), or action self-attention plateaus (AutoHorizon~\cite{autohorizon2026}) improve nominal execution at higher query rates, yet all three remain below $58\%$ completion on high-resistance joints with $23.3\%$ to $26.7\%$ safety overloads because internal model statistics prior to breakaway contain zero friction signal. Relative to holding fixed $6$-step probes without withdrawal ($85.8\%$ low / $79.2\%$ high-resistance completion, but $32.5\%$ safety overloads at $163.4\,\mathrm{N}$ peak), online post-deadband sequential probing with withdrawal induces a deliberate safety--completion trade-off: conservatively aborting borderline high-force pushes trades $5.8$--$10.0$ points of nominal completion ($80.0\%$ / $69.2\%$) for a $6.5\times$ reduction in backstop overloads ($5.0\%$, $109.1\,\mathrm{N}$ peak), while lifting $200\,\mathrm{ms}$ delayed high-resistance completion from $46.0\%$ to $72.0\%$ when paired with Eq.~\eqref{eq:floquet-blend} (Appendix~\ref{app:ext-probing}). Once probing exposes free versus jammed motion, the policy must branch across that feedback instant.

\section{Chunk Supervision Across Post-Contact Bifurcations}
\label{sec:supervision}

\begin{figure*}[t]
\centering
\includegraphics[width=0.92\textwidth]{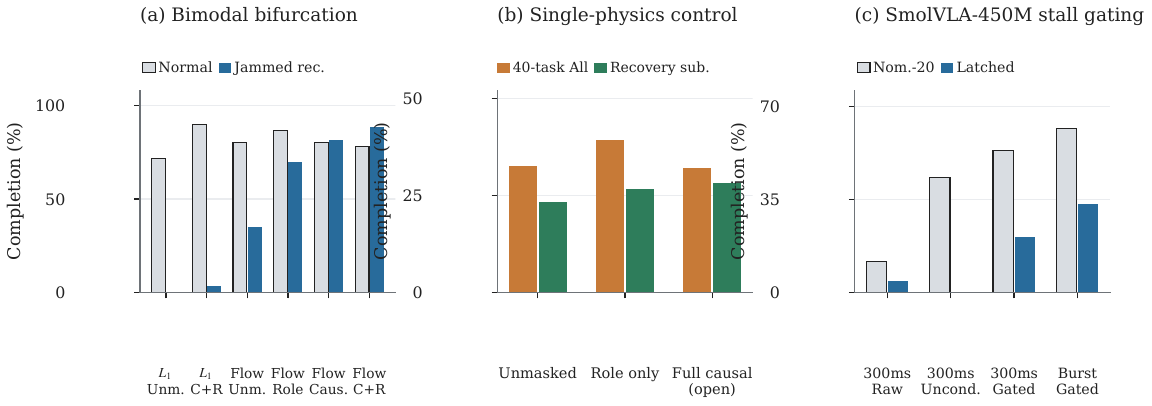}
\caption{Causal boundary masking across post-contact bifurcations, single-physics negative control, and pretrained SmolVLA-450M stall gating. (a)~On bimodal contact bifurcations, unimodal $L_1$ regression collapses on jammed trials ($3.3\%$), whereas conditional flow matching improves monotonically with role filtering, causal truncation, and contact replanning ($35.0\% \to 88.3\%$). (b)~On $40$ single-physics \textsc{LIBERO-SelfTest} tasks without mid-chunk branching, open-loop truncation offers no gain over role filtering ($32.1\%$ vs.\ $39.2\%$). (c)~Under $300\,\mathrm{ms}$ stalled lag on pretrained SmolVLA-450M, unconditional prefix skipping discards unlatch motions ($0.0\%$ latched), whereas displacement-gated prefix alignment restores latched recovery.}
\label{fig:bifurcation}
\vspace{-4pt}
\end{figure*}

\subsection{Post-Contact Median Collapse and Causal Boundary Masking}
Learning recovery behavior from multi-step demonstrations introduces a supervision boundary whenever a sliding $H$-step training window straddles a contact-feedback instant $t_{\mathrm{fb}}(t)$ where latent resistance is revealed. Prior to $t_{\mathrm{fb}}(t)$, normal and jammed joints share visually identical approach histories; after $t_{\mathrm{fb}}(t)$, normal joints require continued forward pushing $a^{\mathrm{low}}$ (probability $1 - p$), whereas jammed joints require withdrawal and unlatching $a^{\mathrm{high}}$ (probability $p < 0.5$). If a single pre-contact chunk is trained via unmasked coordinate-wise $L_1$ regression to predict steps past $t_{\mathrm{fb}}(t)$, its expected post-boundary risk is uniquely minimized by the marginal median (Proposition~3 in Appendix~\ref{app:prop-median}):
\begin{equation}
\begin{aligned}
\mathcal{R}_{L_1}(\hat{a}) &= (1 - p)\bigl\|\hat{a} - a^{\mathrm{low}}\bigr\|_1 + p\bigl\|\hat{a} - a^{\mathrm{high}}\bigr\|_1 \\
&\Longrightarrow\;\; \hat{a}^{\star} \equiv a^{\mathrm{low}} \quad \forall\, p < 0.5,
\end{aligned}
\label{eq:median-collapse}
\end{equation}
collapsing predictions strictly onto the majority forward-push branch regardless of how sharply the jammed recovery action diverges.

In addition, failure-and-recovery demonstrations often begin with induced failure-setup motions that should remain in observation context without being imitated. Combining a binary role mask $m_t^{\mathrm{role}} \in \{0, 1\}$ (excluding induced failure-setup steps) with causal truncation at the next contact-feedback boundary $t_{\mathrm{fb}}(t)$ yields the boundary-masked chunk objective:
\begin{equation}
\mathcal{L}_{\mathrm{act}}(t) = \sum_{j=0}^{H-1} m_{t+j}^{\mathrm{role}} \, \mathbf{1}\bigl[t + j < t_{\mathrm{fb}}(t)\bigr] \, \ell\bigl(\hat{a}_{t,j}, a_{t+j}^{\star}\bigr),
\label{eq:causal-mask}
\end{equation}
where $\ell$ is either $L_1$ regression or conditional flow-matching velocity loss~\cite{lipman2023flow}. Because Eq.~\eqref{eq:causal-mask} stops supervising actions beyond $t_{\mathrm{fb}}(t)$ before contact feedback is observed, training-time truncation is paired at deployment with stall-triggered replanning so a fresh post-contact chunk is queried once contact separates the two branches---detected via contact force ($>\!45\,\mathrm{N}$) and sub-millimeter progress ($<\!0.8\,\mathrm{mm}$) in simulation, or directly via proportional operational-space tracking residual $\|x_{\mathrm{cmd}}-x\|_2 \gtrsim F/K_p$ plus $<\!0.8\,\mathrm{mm}$ progress on sensorless arms (Appendix~\ref{app:protocols}).

\subsection{Bimodal vs.\ Single-Physics Verification on \textsc{LIBERO-SelfTest}}

\begin{table}[t]
\centering
\caption{Ablation of Causal Boundary Masking Across Bimodal Contact and Single-Physics Regimes on \textsc{LIBERO-SelfTest} (\%).}
\label{tab:takeover}
\scriptsize
\setlength{\tabcolsep}{3.5pt}
\begin{tabular}{lccccc}
\toprule
\multicolumn{6}{l}{\emph{A. Bimodal contact bifurcation ($n = 120$/row, $180\,\mathrm{N}$ force limit)}} \\
Head / Supervision & Norm. & JamRec & All & Ovld$\downarrow$ & $F_{\mathrm{pk}}$ ($\mathrm{N}$) \\
\midrule
$L_1$ Unmasked (open) & 71.7 & 0.0 & 35.8 & 14.2 & 200.3 \\
$L_1$ Role-only (open) & 80.0 & 48.3 & 64.2 & 11.7 & 185.6 \\
$L_1$ Causal (open) & 88.3 & 1.7 & 45.0 & 7.5 & 145.3 \\
$L_1$ Causal + Replan & \textbf{90.0} & 3.3 & 46.7 & \textbf{5.8} & \textbf{136.7} \\
\midrule
Flow Unmasked (open) & 80.0 & 35.0 & 57.5 & 15.8 & 185.1 \\
Flow Role-only (open) & 86.7 & 70.0 & 78.3 & 20.0 & 194.3 \\
Flow Causal (open) & 80.0 & 81.7 & 80.8 & 19.2 & 173.4 \\
\textbf{Flow Causal + Replan} & 78.3 & \textbf{88.3} & \textbf{83.3} & 16.7 & 190.3 \\
\midrule
\multicolumn{6}{l}{\emph{B. Single-physics $40$-task \textsc{LIBERO-SelfTest} (TaskFiLM-ResNet18, no replan)}} \\
Mask Objective & Spat. & Obj. & Goal & All & Rec. \\
\midrule
Unmasked ($m_{t,j} \equiv 1$) & 46.7 & 40.0 & 31.7 & 32.5 & 23.3 \\
Role only ($m_{t+j}^{\mathrm{role}}$) & \textbf{51.7} & \textbf{51.7} & \textbf{35.0} & \textbf{39.2} & 26.7 \\
Full causal (open) & 46.7 & 43.3 & 28.3 & 32.1 & \textbf{28.3} \\
\bottomrule
\end{tabular}
\vspace{2pt}

\parbox{\linewidth}{\scriptsize\emph{Note:} $3$-seed means of Eq.~\eqref{eq:causal-mask}. Panel~A: $960$ MuJoCo rollouts (L1- vs.\ Flow-ChunkNet); Panel~B: $720$ open-loop rollouts.}
\end{table}

To test whether causal chunk truncation helps universally or specifically across post-contact bifurcations, we construct \textsc{LIBERO-SelfTest} across all $130$ LIBERO tasks, comprising $1{,}040$ replay-verified trajectories ($532$ nominal completions and $508$ failure-and-recovery episodes with $2{,}848$ annotated probe and feedback boundaries and $327{,}637$ eight-step windows; suite breakdown and protocols in Appendix~\ref{app:protocols}). We evaluate matched deterministic regression and conditional flow-matching policies across two regimes in Table~\ref{tab:takeover} and Fig.~\ref{fig:bifurcation}a--b: articulated joints with bimodal post-contact mechanics (normal push vs.\ jammed unlatch-withdrawal under a $180\,\mathrm{N}$ safety limit) and $40$ standard single-physics tasks where mechanics never branch mid-chunk.

The contrast between the two regimes isolates the exact scope of boundary truncation:
\begin{itemize}
    \item \emph{Bimodal post-contact bifurcations (Table~\ref{tab:takeover}A):} Conditional flow matching improves monotonically as each boundary mechanism is added---role filtering lifts overall completion from $57.5\%$ to $78.3\%$ (jammed recovery from $35.0\%$ to $70.0\%$), causal truncation at $t_{\mathrm{fb}}(t)$ raises jammed recovery to $81.7\%$, and pairing truncation with contact-triggered replanning reaches $88.3\%$ jammed recovery and $83.3\%$ overall completion. By contrast, unimodal $L_1$ regression reaches $90.0\%$ on normal joints even with truncation and replanning, yet collapses onto forward pushing on jammed trials ($3.3\%$ recovery), confirming Eq.~\eqref{eq:median-collapse}.
    \item \emph{Single-physics negative control (Table~\ref{tab:takeover}B):} Across $40$ tasks without mid-chunk physical branching or event-triggered replanning, excluding failure-setup prefixes lifts mean completion from $32.5\%$ to $39.2\%$, whereas open-loop causal truncation sacrifices chunk-tail continuity and lowers overall completion to $32.1\%$. Truncating supervision is therefore not a generic substitute for full horizons; it resolves multi-branch contact bifurcations specifically when paired with multimodal action heads and event-triggered replanning.
\end{itemize}

\subsection{Scaling to Pretrained Flow VLAs Under Asynchronous Stall}

\begin{table}[t]
\centering
\caption{Post-Contact Bifurcation Supervision and Delayed Stall Gating on Pretrained SmolVLA-450M ($26{,}744$ Rollouts, $3$ Seeds).}
\label{tab:smolvla-bifurc}
\scriptsize
\setlength{\tabcolsep}{1.8pt}
\begin{tabular}{lcccc}
\toprule
\multicolumn{5}{l}{\emph{A. Undelayed supervision \& subspace WiSE-FT ($19{,}232$ rollouts)}} \\
Supervision / Schedule & Nom.-40 & Norm. & Latched & All (\%) \\
\midrule
Base + PACE / AutoHor.~\cite{pace2026} & 70.0 / \textbf{81.2} & \textbf{93.8} & 0.0 & 46.9 \\
$L_1$ Unmasked + PACE & $70.4 \pm 2.4$ & 91.7 & $0.0 \pm 0.0$ & 45.8 \\
Flow Unmasked (open-$24$) & $57.1 \pm 1.6$ & 55.6 & $19.4 \pm 6.4$ & 37.5 \\
Flow Causal + Event Replan & $60.8 \pm 1.2$ & 72.9 & $32.6 \pm 2.6$ & 52.8 \\
Uniform WiSE-FT ($\beta{=}0.75$) & $56.3 \pm 2.0$ & 67.6 & $41.7 \pm 2.3$ & 54.6 \\
\textbf{Subspace $(0.35, 0.75)$ + PACE} & $\mathbf{62.9 \pm 2.4}$ & 72.2 & $41.7 \pm 8.2$ & \textbf{56.9} \\
\textbf{Subspace $(0.50, 0.85)$ + PACE} & $62.5 \pm 3.7$ & 65.7 & $\mathbf{47.2 \pm 9.9}$ & 56.5 \\
\midrule
\multicolumn{5}{l}{\emph{B. Delayed bifurcations ($7{,}512$ rollouts; col.\ 2: Nom.-20)}} \\
$200\,\mathrm{ms}$ Raw / Gated Fl.+Pref. & 23.3 / \textbf{61.7} & 38.9 / \textbf{72.2} & 19.4 / \textbf{26.4} & 29.2 / \textbf{47.9} \\
$300\,\mathrm{ms}$ Raw / Uncond.\ Fl. & 11.7 / 43.3 & 15.3 / 51.4 & 4.2 / 0.0 & 9.7 / 25.7 \\
\textbf{$300\,\mathrm{ms}$ Gated Fl.+Pref.} & \textbf{53.3} & \textbf{69.4} & \textbf{20.8} & \textbf{45.1} \\
\textbf{Burst $\{2,7\}$ Gated Fl.+Pref.} & \textbf{61.7} & \textbf{62.5} & \textbf{33.3} & \textbf{47.9} \\
\bottomrule
\end{tabular}
\vspace{2pt}

\parbox{\linewidth}{\scriptsize\emph{Note:} Fine-tuned rows report $3$-seed means under replay and subspace WiSE-FT~\cite{wiseft2022} $(\beta_{\mathrm{h}}, \beta_{\mathrm{e}})$. Panel~B gates prefix skipping on realized in-flight displacement $\|x(t_{\mathrm{arr}}) - x(t_{\mathrm{obs}})\|_2 \ge 4.5\,\mathrm{mm}$.}
\end{table}

Extending this audit to pretrained SmolVLA-450M~\cite{smolvla2025} across $12$ post-contact latch-bifurcation tasks co-trained with replay on $28$ nominal tasks ($26{,}744$ rollouts; Table~\ref{tab:smolvla-bifurc}, Fig.~\ref{fig:bifurcation}c) confirms the same boundary behavior at foundation-model scale. Both base SmolVLA with adaptive horizons (PACE, AutoHorizon) and unmasked $L_1$ fine-tuning score $0.0\%$ on latched episodes despite exceeding $91\%$ on unlatched joints. Under conditional flow matching, causal truncation paired with event replanning and subspace weight interpolation~\cite{wiseft2022} (restraining projection-head drift while adapting transformer blocks) restores latched recovery to $47.2\%$ while preserving $40$-task nominal retention.

Evaluating asynchronous $200$--$300\,\mathrm{ms}$ inference delay alongside latched stalls further exposes a compound boundary failure in standard prefix skipping (RTC~\cite{rtc2025}). When a latched joint immobilizes the gripper during $300\,\mathrm{ms}$ inference, unconditionally advancing to a later chunk index discards the initial unlatch motion and drops latched recovery to $0.0\%$ (Table~\ref{tab:smolvla-bifurc}B). Using only RGB and proprioceptive state (no force input), gating prefix skipping on realized in-flight displacement ($\ge\!4.5\,\mathrm{mm}$, set just above Section~\ref{sec:probing}'s $\approx\!2\,\mathrm{mm}$ finger-pad compliance deadband so elastic pad compression under stall is not mistaken for free-space travel) retains index $0$ when stalled and restores $20.8\%$ and $33.3\%$ latched completion under $300\,\mathrm{ms}$ lag and burst jitter. Even when imitation preserves recovery branches, however, reinforcement post-training can still erase multi-route flexibility.

\section{Opening-Branch Narrowing and Retry Recovery in RL Post-Training}
\label{sec:rl_narrowing}

\begin{figure*}[t]
\centering
\includegraphics[width=0.92\textwidth]{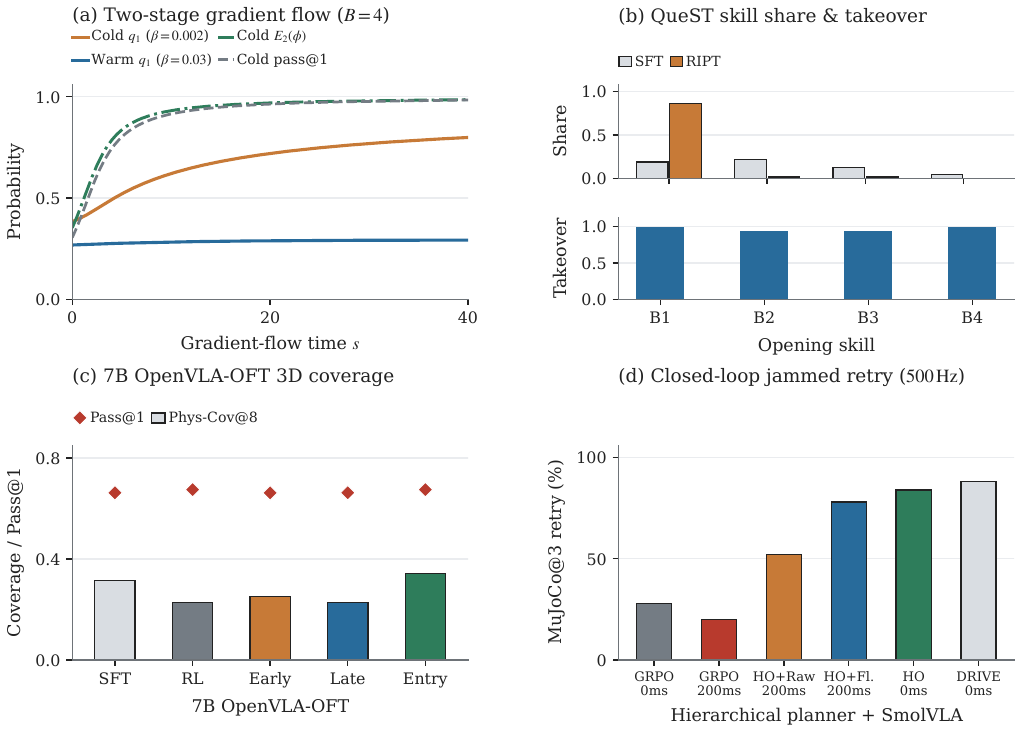}
\caption{Two-stage opening narrowing and closed-loop retry recovery across VLA post-training. (a)~Two-stage KL-regularized gradient flow (Eq.~\eqref{eq:coupled-ode-system}) comparing cold-start collapse against warm-start retention and intact unvisited continuation $E_2(\phi)$. (b)~QueST opening-skill share (top) and interventional RIPT takeover completion (bottom) across dominant skill B1 and suppressed skills B2--B4. (c)~7B OpenVLA-OFT $15\,\mathrm{mm}$ $3\mathrm{D}$ approach-corridor coverage (Phys-Cov@8, bars) and Pass@1 (diamonds) across SFT, RL, LoRA interpolation, and $40$-step opening handoff. (d)~$500\,\mathrm{Hz}$ MuJoCo $3$-attempt jammed-subgoal retry comparing vanilla GRPO, opening handoff (HO) with raw vs.\ Floquet-damped SmolVLA, and DRIVE-GRPO~\cite{drive2026}.}
\label{fig:rl-narrowing}
\vspace{-4pt}
\end{figure*}

\subsection{Two-Stage Policy-Gradient Localization}
Online outcome reinforcement learning (RIPT-VLA~\cite{ript2025}, SimpleVLA-RL~\cite{simplevlarl2026}) sharpens single-rollout success across visuomotor VLAs and symbolic planners, yet contracts behavioral breadth~\cite{drive2026,extoken2026} and frequently leaves multi-attempt retries repeating nearly identical failed trajectories~\cite{interactvla2026}. To locate where diversity is lost, consider a two-stage policy that first selects an opening branch $b \in \{1, \dots, B\}$ (such as a $3\mathrm{D}$ approach corridor or initial subgoal) with softmax probabilities $q_b(z)$ relative to pre-RL prior $q_b^0$, and then executes shared continuation parameters $\phi$ with conditional success rate $E_b(\phi)$. Under KL-regularized policy-gradient flow, opening logits $z_b$ and continuation parameters $\phi$ evolve as (Proposition~4 in Appendix~\ref{app:prop-rl}):
\begin{equation}
\begin{aligned}
\frac{dz_b}{ds} &= \kappa_z q_b \left[ \bigl(E_b(\phi) - \bar{E}\bigr) - \beta \left(\ln\frac{q_b}{q_b^0} - D_{\mathrm{KL}}(q \,\|\, q^0)\right) \right], \\
\frac{d\phi}{ds} &= \kappa_\phi \sum_{b=1}^{B} q_b \nabla_\phi E_b(\phi).
\end{aligned}
\label{eq:coupled-ode-system}
\end{equation}
Equation~\eqref{eq:coupled-ode-system} reveals a sharp structural asymmetry between the opening boundary and downstream execution (Fig.~\ref{fig:rl-narrowing}a). Under weak KL regularization ($\beta \to 0$), any initial return gap drives $q_b$ exponentially onto the single highest-return opening branch; yet whenever continuation skills share positive gradient alignment across branches, downstream execution $E_b(\phi)$ inside suppressed branches remains intact and continues to improve. Diversity collapse is therefore localized at the initial commitment boundary rather than within branch execution, predicting that running the pre-RL policy solely across the opening segment before handing control to the RL policy can recover route diversity without retraining or accuracy loss.

\subsection{Localization Audit Across 7B VLA and Enumerable Reasoning}

\begin{table*}[t]
\centering
\caption{Opening-Boundary Recovery on 7B OpenVLA-OFT, Closed-Loop $500\,\mathrm{Hz}$ MuJoCo Jammed Retry, and 7.65B SimpleVLA-RL Robustness.}
\label{tab:rl-retry}
\scriptsize
\begin{minipage}[t]{0.33\textwidth}
\centering
\emph{(a) 7B OpenVLA-OFT ($400$ rollouts)}\\[2pt]
\setlength{\tabcolsep}{2.6pt}
\begin{tabular}{lccc}
\toprule
Condition ($k = 8$) & P@1 & P@8 & Cov@8 \\
\midrule
SFT (base) & 0.662 & 0.700 & 0.314 \\
7B RL (ctrl) & \textbf{0.675} & 0.700 & 0.227 \\
Early L0--11 & 0.662 & 0.700 & 0.252 \\
Mid-late L16--31 & 0.662 & 0.700 & 0.228 \\
\textbf{Opening SFT$\to$RL} & \textbf{0.675} & 0.700 & \textbf{0.342} \\
\midrule
QueST-RIPT ($20$t) & \multicolumn{3}{c}{$\mathcal H: 2.48{\to}1.65$, tk.\ $0.902$} \\
\bottomrule
\end{tabular}
\end{minipage}\hfill
\begin{minipage}[t]{0.34\textwidth}
\centering
\emph{(b) Blocked-Subgoal Retry ($150$ tasks)}\\[2pt]
\setlength{\tabcolsep}{2.3pt}
\begin{tabular}{lccc}
\toprule
Planner ($k=32$) & P@1 & R@3 & MuJ@3 \\
\midrule
Multi-branch SFT & 0.678 & 0.618 & 70.0 \\
Vanilla GRPO & \textbf{0.990} & 0.236 & 28.0 \\
Entropy-GRPO & 0.979 & 0.301 & 30.0 \\
DRIVE-GRPO~\cite{drive2026} & 0.949 & \textbf{0.819} & \textbf{88.0} \\
\textbf{Opening SFT$\to$GRPO} & 0.909 & 0.710 & 84.0 \\
\cmidrule(lr){1-4}
$200\,\mathrm{ms}$ GRPO+Raw & 0.990 & 0.236 & 20.0 \\
\textbf{$200\,\mathrm{ms}$ HO+Raw/Fl.} & 0.903 & 0.710 & 52.0/\textbf{78.0} \\
\bottomrule
\end{tabular}
\end{minipage}\hfill
\begin{minipage}[t]{0.31\textwidth}
\centering
\emph{(c) 7.65B SimpleVLA-RL ($M=1$)}\\[2pt]
\setlength{\tabcolsep}{2.2pt}
\begin{tabular}{lccc}
\toprule
Policy Condition & R@1 & R@6 & Init-P. \\
\midrule
SFT ($M=1$) & 2.5 & 10.0 & 5.0 \\
SimpleVLA-RL & 22.5 & 40.0 & 20.0 \\
\textbf{Early L0--11} & 20.0 & 35.0 & \textbf{45.0} \\
Mid-late L16--31 & 27.5 & \textbf{45.0} & 25.0 \\
Opening SFT$\to$RL & \textbf{28.3} & 40.0 & 20.0 \\
\cmidrule(lr){1-4}
RIPT RL / \textbf{HO} & \multicolumn{3}{c}{Sprd $17.6{\to}\mathbf{19.2}$, P@6 $96.7$} \\
\bottomrule
\end{tabular}
\end{minipage}
\vspace{2pt}

\parbox{\linewidth}{\scriptsize\emph{Note:} (a)~RIPT-VLA~\cite{ript2025} on $50$ perturbed LIBERO-Spatial states ($15\,\mathrm{mm}$ corridor Phys-Cov@8; $40$-step opening handoff). (b)~Qwen3-1.7B planner paired with SmolVLA-450M in $500\,\mathrm{Hz}$ MuJoCo ($450$ multi-attempt episodes; HO denotes opening handoff). (c)~SimpleVLA-RL~\cite{simplevlarl2026} on LIBERO-10 and LIBERO-Plus~\cite{liberoplus2026} initial-pose shifts (Init-P.). Full enumerable \emph{Countdown} audit ($135$ problems) in Appendix~\ref{app:countdown}.}
\end{table*}

Experiments across discrete-skill QueST~\cite{quest2024}, 7B OpenVLA-OFT~\cite{oft2025}, and exhaustively enumerable \emph{Countdown} reasoning~\cite{pan2025countdown} confirm Eq.~\eqref{eq:coupled-ode-system} (Table~\ref{tab:rl-retry}a, Fig.~\ref{fig:rl-narrowing}b--c). On QueST across $20$ tasks, RIPT post-training contracts opening-skill entropy from $2.48$ to $1.65\,\mathrm{nats}$, yet clamping suppressed opening skills (B2--B4) at the initial step before handing control to the RL policy maintains $0.902$ closed-loop completion. On 7B OpenVLA-OFT across $400$ perturbed LIBERO-Spatial rollouts, RIPT raises single-sample accuracy from $0.662$ to $0.675$ while shrinking $3\mathrm{D}$ approach-corridor coverage from $0.314$ to $0.227$. Whereas static LoRA interpolation yields modest shifts ($0.228$ to $0.252$), running the pre-RL policy across the opening $40$ steps ($5$ chunks spanning pre-grasp corridor commitment on LIBERO-Spatial, generalizable event-wise until initial grasp deceleration on variable-horizon tasks) before handing off to RL restores $3\mathrm{D}$ corridor coverage to $0.342$ at full accuracy ($0.675$). Because continuous $3\mathrm{D}$ workspaces cannot exhaustively enumerate unvisited valid trajectories to verify $E_b(\phi)$ once $q_b \to 0$, we pair this VLA audit with $135$ held-out \emph{Countdown} problems~\cite{pan2025countdown} whose complete ground-truth solution trees are exact and enumerable, confirming identical entrance localization under full-parameter PPO and GRPO ($0.203 \to 0.289$ canonical coverage; Appendix~\ref{app:countdown}).

\subsection{Closed-Loop Jammed Retry and Perturbation Robustness}
Finally, we verify that restoring opening-boundary breadth directly resolves multi-attempt manipulation retries when dominant physical routes are blocked (Table~\ref{tab:rl-retry}b--c, Fig.~\ref{fig:rl-narrowing}d). In $500\,\mathrm{Hz}$ MuJoCo hierarchical execution pairing a GRPO-trained symbolic planner with SmolVLA-450M across $150$ subgoal tasks, jamming the primary physical subgoal causes vanilla GRPO to repeat the blocked route, stalling $3$-attempt completion at $28.0\%$ undelayed and $20.0\%$ under $200\,\mathrm{ms}$ lag. Without any diversity-regularized RL retraining (DRIVE~\cite{drive2026}), test-time opening handoff raises $3$-attempt symbolic recovery from $0.236$ to $0.710$, and pairing opening handoff with Floquet-damped prefix control (Eq.~\eqref{eq:floquet-blend}) lifts $200\,\mathrm{ms}$ closed-loop MuJoCo retry completion from $52.0\%$ to $78.0\%$. Conversely, Table~\ref{tab:rl-retry}c exposes the prior-support boundary of test-time handoff: when SFT demonstrations cover only one trajectory corridor ($M=1$ on $7.65\mathrm{B}$ SimpleVLA-RL~\cite{simplevlarl2026}), \textsc{Opening SFT$\to$RL} lifts nominal \textsc{R@1} ($22.5\% \to 28.3\%$) but leaves initial-pose shift robustness (\textsc{Init-P.}) at $20.0\%$ because the pre-RL prior contains no alternate route, whereas attenuating early visuomotor blocks (\textsc{Early L0--11}) relaxes spatial feature sharpening to reach $45.0\%$ on LIBERO-Plus~\cite{liberoplus2026}.

\section{Discussion and Limitations}
\label{sec:conclusion}

Across physical execution, imitation learning, and reinforcement post-training, multi-step commitment succeeds inside smooth regimes because closed-loop tracking and learned continuation skills are locally contractive, and breaks specifically where physical mechanics jump or future decisions branch. Separating smooth interiors from sharp boundaries avoids a pervasive deployment tradeoff: policies need not sacrifice open-loop throughput in free space, retrain with auxiliary world models, or trade away single-try RL precision to preserve contact safety, jam recovery, and retry diversity. Instead, lightweight interventions targeted strictly at contact entry, post-deadband feedback, and opening branch selection suffice to repair failure modes that uniform horizon tuning leaves unresolved.

Several limitations point to concrete directions for future work. First, our closed-loop physical audits are conducted in $500\,\mathrm{Hz}$ MuJoCo contact simulation; transferring post-deadband sequential probing and closure withdrawal to physical hardware must contend with unmodeled transmission backlash, non-planar handle compliance, and noisier state estimation, motivating integration with wrist force-torque or tactile sensing when proprioceptive displacement alone has low signal-to-noise ratio. Second, our causal chunk mask targets discrete contact-feedback instants where mechanics bifurcate sharply; tasks involving deformable cloth, granular media, or continuous wiping exhibit gradual regime shifts that may benefit from soft boundary weighting rather than hard truncation. Finally, test-time opening handoff recovers route breadth already present in the pre-RL policy; when demonstrations cover only a single valid corridor ($M = 1$), discovering new routes still requires exploration-promoting training objectives such as DRIVE~\cite{drive2026} alongside early-layer update attenuation.

\IEEEtriggeratref{35}
\bibliographystyle{IEEEtran}
\bibliography{references}

\clearpage
\appendices
\makeatletter\setlength{\@fptop}{0pt}\makeatother

\section{Mathematical Proofs and Mechanics Derivations}
\label{app:proofs-control}
\label{app:proofs-learning}

\subsection{Delayed Reference Decomposition and Sawtooth Floquet Stability}
\label{app:prop-delay}

\noindent\textbf{Proposition 1 (Delayed Target Bias and Periodic Sawtooth Contraction).}
\emph{Let $x^\star, g^\star \in C^2$ denote a smooth nominal end-effector trajectory and Operational Space Control (OSC) target sampled every $\Delta t$ under a base lag of $d_0$ steps and executed chunk stride $c$. First, anchoring a lookahead increment of index $i \in \{0,\dots,d\}$ to a historical pose $x^\star(t_{n-m})$ incurs a first-order bias that vanishes during steady transit for $(i,m)=(0,0)$ and during terminal docking for $(i,m)=(0,d)$. Second, under periodic observation age $d_n = d_0 + (n \bmod c)$ and closed-loop step contraction $\eta_c \in (0,1]$, linear tracking error decays exponentially across chunks if and only if the Floquet contraction factor in Eq.~\eqref{eq:floquet-blend} satisfies $\bar{\rho}(d_0, c, \gamma) < 1$.}

\smallskip
\noindent\emph{Proof.}
Expanding the anchored target $g_n^{(i,m)}$ around $t_n$ to first order under zero-mean open-loop prediction variance $\sigma_{\mathrm{pred}}^2(i)$ separates steady-motion velocity drift from terminal deceleration overshoot:
\begin{equation}
\begin{aligned}
\mathbb{E}\bigl[\|g_n^{(i,m)} - g^\star(t_n)\|_2^2\bigr] &= \bigl\|(d - i - m)\Delta t\,\dot{x}^\star(t_n) \\
&\quad + (i - d)\Delta t\,\dot{g}^\star(t_n)\bigr\|_2^2 + \sigma_{\mathrm{pred}}^2(i).
\end{aligned}
\label{eq:app-bias-var}
\end{equation}
At minimum open-loop variance ($i = 0$), steady transit ($\dot{x}^\star = \dot{g}^\star$) cancels bias under current-pose anchoring ($m = 0$), whereas terminal docking ($\dot{g}^\star = 0$) cancels forward overshoot under enqueue anchoring ($m = d$). Penalizing unlagged overshoot scale $\|v_n^{\mathrm{obs}}\|_2$ against mean commanded chunk speed $\|v_n^{\mathrm{cmd}}\|_2$ through a convex quadratic surrogate yields the velocity-ratio interpolation weight in Eq.~\eqref{eq:floquet-blend}:
\begin{equation}
\begin{aligned}
w_n &= \operatorname*{arg\,min}_{w \in [0,1]} \Bigl[(1 - w)^2 \|v_n^{\mathrm{obs}}\|_2^2 + w^2 \|v_n^{\mathrm{cmd}}\|_2^2\Bigr] \\
&= \frac{\|v_n^{\mathrm{obs}}\|_2^2}{\|v_n^{\mathrm{obs}}\|_2^2 + \|v_n^{\mathrm{cmd}}\|_2^2 + \varepsilon}.
\end{aligned}
\label{eq:app-blend-weight}
\end{equation}
Over a chunk cycle of lifted history horizon $D = d_0 + c$, let $e_k \in \mathbb{R}^D$ denote the $k$-th standard basis vector and $\mathbf{S}_{D-1} = [\mathbf{I}_{D-1},\, \mathbf{0}]$ denote the one-step history shift operator. Each delayed scalar tracking coordinate evolves under the sparse companion operator
\begin{equation}
\mathbf{A}_{d_n}(\gamma) =
\begin{bmatrix}
\bigl(e_1 - \gamma\eta_c\, e_{d_n+1}\bigr)^\top \\[2pt]
\mathbf{S}_{D-1}
\end{bmatrix} \in \mathbb{R}^{D \times D}.
\label{eq:app-companion}
\end{equation}
By Floquet--Lyapunov theory~\cite{floquet1883equations}, composing $\mathbf{A}_{d_0+k}(\gamma)$ over one full stride is exponentially contractive if and only if $\bar{\rho}(d_0, c, \gamma) < 1$. In the stationary single-step limit ($c = 1$), coalescing positive real roots and unit-circle crossings of the characteristic polynomial $\lambda^{d+1} - \lambda^d + \gamma\eta_c = 0$ give the monotonic-damping and oscillatory stability boundaries overlaid in Fig.~\ref{fig:approach-contact}a:
\begin{equation}
\gamma_{\mathrm{mono}}(d) = \frac{d^d}{(d+1)^{d+1}\eta_c}, \qquad \gamma_{\mathrm{crit}}(d) = \frac{2\sin\!\bigl(\frac{\pi}{2(2d+1)}\bigr)}{\eta_c}.
\label{eq:app-stability-thresholds}
\end{equation}
\null\hfill$\blacksquare$

\subsection{Zero-Information Pad Deadband and Sequential Probing Horizon}
\label{app:prop-deadband}

\noindent\textbf{Proposition 2 (Compliance Deadband Fisher Information and Expected SPRT Stopping Time).}
\emph{Suppose an articulated joint with unknown breakaway threshold $\theta \ge \theta_{\min} > 0$ is contacted through an elastic finger pad of stiffness $k_{\mathrm{pad}}$ under increasing actuation force $f_{\mathrm{act}}(t)$. Throughout the pre-breakaway compression window ($j \le d_{\mathrm{dead}}$), displacement observations carry strictly zero Fisher information about $\theta$. Past breakaway, sequential log-likelihood testing (Eq.~\eqref{eq:wald-sprt}) under stationary post-deadband drift $\bar{\delta}(\theta) \neq 0$, increment variance $\bar{v} > 0$, and symmetric decision bound $B$ obeys Eq.~\eqref{eq:app-sprt-bound}.}

\smallskip
\noindent\emph{Proof.}
While applied force remains below the minimum static breakaway threshold $\theta_{\min}$, the rigid joint stays immobilized at its initial coordinate $q_0$ and all end-effector motion is absorbed by elastic pad deformation:
\begin{equation}
\begin{aligned}
x_{\mathrm{ee}}(t_j) - q_0 &= \frac{f_{\mathrm{act}}(t_j)}{k_{\mathrm{pad}}} \quad (j \le d_{\mathrm{dead}}) \\
&\Longrightarrow\;\; \frac{\partial \mu_j(\theta)}{\partial \theta} = \mathbf{0}, \quad \mathcal{I}_m(\theta) = 0.
\end{aligned}
\label{eq:app-deadband-fisher}
\end{equation}
Because the mean displacement trajectory $\mu_j(\theta)$ does not depend on $\theta$ prior to breakaway, neither visual frames nor internal policy statistics carry friction signal during the deadband. Once breakaway occurs ($j > d_{\mathrm{dead}}$), applying optional stopping to the exponential martingale of post-deadband log-likelihood increments and neglecting boundary overshoot at $\pm B$ yields, via Wald's identity~\cite{wald1945sequential}, the expected probe stopping horizon:
\begin{equation}
\mathbb{E}_\theta[N_{\mathrm{SPRT}}] \approx d_{\mathrm{dead}} + \frac{B}{\bar{\delta}(\theta)} \tanh\!\left(\frac{B\,\bar{\delta}(\theta)}{\bar{v}}\right).
\label{eq:app-sprt-bound}
\end{equation}
\null\hfill$\blacksquare$

\subsection{Conditional-Median Collapse Across Post-Contact Bifurcations}
\label{app:prop-median}

\noindent\textbf{Proposition 3 (Majority-Branch Collapse Under Unmasked $L_1$ Regression).}
\emph{Consider a post-boundary action coordinate taking normal continuation value $a^{\mathrm{low}}$ with probability $1 - p$ and jammed withdrawal value $a^{\mathrm{high}} < a^{\mathrm{low}}$ with probability $p \in (0, 1/2)$ given an identical pre-contact observation $\mathcal{O}_t$. Minimizing unmasked coordinate-wise $L_1$ error across $t_{\mathrm{fb}}(t)$ collapses uniquely onto $\hat{a}^\star = a^{\mathrm{low}}$, independent of branch separation $|a^{\mathrm{low}} - a^{\mathrm{high}}|$.}

\smallskip
\noindent\emph{Proof.}
Conditioned on pre-contact observation $\mathcal{O}_t$, the piecewise-linear $L_1$ risk in Eq.~\eqref{eq:median-collapse} admits the subdifferential
\begin{equation}
\partial \mathcal{R}_{L_1}(\hat{a}) =
\begin{cases}
\{2p - 1\}, & a^{\mathrm{high}} < \hat{a} < a^{\mathrm{low}}, \\
[2p - 1,\; 1], & \hat{a} = a^{\mathrm{low}}.
\end{cases}
\label{eq:app-subgrad}
\end{equation}
Whenever jammed episodes form the minority regime ($p < 1/2$), the interior slope $2p - 1$ is strictly negative while zero lies strictly inside the subdifferential interval at $a^{\mathrm{low}}$, establishing $\hat{a}^\star = a^{\mathrm{low}}$ as the unique global minimizer. Applying causal boundary masking (Eq.~\eqref{eq:causal-mask}) removes unconditioned post-boundary supervision prior to touch; once contact feedback arrives at $t_{\mathrm{fb}}(t)$ and triggers replanning, post-contact state observations separate the two regimes so conditional flow matching~\cite{lipman2023flow} transports samples cleanly onto the unlatch-withdrawal mode. \null\hfill$\blacksquare$

\subsection{Opening-Branch Collapse and Shared Continuation Flow in RL}
\label{app:prop-rl}

\noindent\textbf{Proposition 4 (Exponential Opening Extinction Without Continuation Degradation).}
\emph{Consider the two-stage policy in Section~\ref{sec:rl_narrowing} under KL regularization weight $\beta \ge 0$. In unregularized outcome optimization ($\beta = 0$), if dominant opening branch $b = 1$ maintains a strict return margin $\Delta_{\min} > 0$ over every alternative branch $b \ge 2$, then alternative opening probabilities decay exponentially to zero while their downstream continuation success $E_b(\phi)$ strictly increases whenever continuation gradients align positively across branches.}

\smallskip
\noindent\emph{Proof.}
Differentiating expected return minus KL divergence with respect to softmax opening logits $z$ and shared continuation parameters $\phi$ yields the coupled dynamics in Eq.~\eqref{eq:coupled-ode-system}. Setting $\beta = 0$ and integrating the log-probability ratio between the dominant branch and any sub-optimal branch $b \ge 2$ gives exponential opening extinction alongside positive continuation transfer:
\begin{equation}
\begin{aligned}
q_b(s) &\le \frac{q_b(0)}{q_1(0)} \exp\bigl(-\kappa_z \Delta_{\min} s\bigr) \xrightarrow{s \to \infty} 0, \\
\frac{d}{ds} E_b\bigl(\phi(s)\bigr) &= \kappa_\phi \sum_{k=1}^{B} q_k(s)\,\bigl\langle \nabla_\phi E_b(\phi),\, \nabla_\phi E_k(\phi) \bigr\rangle > 0,
\end{aligned}
\label{eq:app-rl-bounds}
\end{equation}
whenever shared motor continuation gradients maintain positive inner-product alignment with the dominant branch. Outcome optimization thus collapses initial branch commitment without degrading downstream motor execution inside unvisited corridors. \null\hfill$\blacksquare$

\subsection{Quasi-Static Curved Self-Locking and Wrist-Yaw Wrench Mechanics}
\label{app:mechanics}

For an axisymmetric object of mass $m$ with smooth convex or flared wall profile $z = \phi(r)$ and local wall inclination $\tan\varphi(r) = |\phi'(r)|$, symmetric horizontal clamping force $f_c$ on opposite rim points induces upward normal reactions and downward Coulomb tangential resistance along each finger pad. Balancing vertical reactions against half-gravity loading per finger during quasi-static lift yields the exact self-locking inequality generalizing Eq.~\eqref{eq:friction-cone}, while wrist-yaw misalignment $\delta\theta_{\mathrm{yaw}}$ at radial offset $\delta r$ across jaw half-span $L_{\mathrm{jaw}}$ generates an asymmetric single-finger vertical wrench:
\begin{equation}
\begin{aligned}
f_c \sin\varphi + \tfrac{1}{2} m g \cos\varphi &\le \mu \bigl(f_c \cos\varphi - \tfrac{1}{2} m g \sin\varphi\bigr)_+, \\
M_z(\delta r, \delta\theta_{\mathrm{yaw}}) &= f_c \bigl(L_{\mathrm{jaw}} \sin\delta\theta_{\mathrm{yaw}} + \delta r \cos\delta\theta_{\mathrm{yaw}}\bigr).
\end{aligned}
\label{eq:app-mechanics}
\end{equation}
Violating the first relation wedges the object upward out of the jaws upon closure, whereas nonzero $M_z$ twists the rim during single-finger touch prior to antipodal engagement.

\section{\textsc{LIBERO-SelfTest} Benchmark and Experimental Protocols}
\label{app:protocols}

\subsection{Simulation Timing, Asynchronous Control, and Event Replanning}
All closed-loop evaluations run in $500\,\mathrm{Hz}$ MuJoCo contact physics ($2\,\mathrm{ms}$ integration step) driven by a $20\,\mathrm{Hz}$ Operational Space Controller ($50\,\mathrm{ms}$ control interval) following a $10$-step neutral-gripper settling phase at episode initialization. Across the delayed execution baselines in Table~\ref{tab:delay}, raw execution applies delayed increments directly to the current end-effector pose; enqueue calibration anchors increments to the historical pose recorded when the observation was captured; temporal ensembling (ACT-TE~\cite{act2023}) averages overlapping FIFO chunk predictions with exponential decay weight $0.01$; prefix alignment~\cite{rtc2025} advances into the predicted chunk by the number of control steps elapsed during inference; and Floquet blending applies Eq.~\eqref{eq:floquet-blend} using analytical stability gains $1.00$ for $100\,\mathrm{ms}$ lag, $0.51$ for $200\,\mathrm{ms}$ lag, and $0.45$ for $300\,\mathrm{ms}$ lag. Under stall-triggered replanning in Section~\ref{sec:supervision}, the open-loop chunk queue flushes and queries a fresh policy chunk whenever measured contact force exceeds $45\,\mathrm{N}$ alongside sub-millimeter progress ($<\!0.8\,\mathrm{mm}$ over two consecutive steps); on low-cost arms without a wrist force-torque sensor, under proportional operational-space stiffness $K_p$ this binary stall trigger maps directly to a Cartesian tracking residual $\|x_{\mathrm{cmd}}-x\|_2 \gtrsim 45\,\mathrm{N}/K_p$ (or joint motor-current residual) paired with $<\!0.8\,\mathrm{mm}$ progress. Episodes exceeding $180\,\mathrm{N}$ are flagged as safety overloads.

\subsection{\textsc{LIBERO-SelfTest} Corpus Construction and Boundary Annotations}
Table~\ref{tab:app-corpus} summarizes the replay-verified \textsc{LIBERO-SelfTest} corpus across all five LIBERO suites ($130$ tasks, $1{,}040$ episodes, and $342{,}059$ executed control steps). Each task pairs nominal pass trajectories with failure-and-recovery episodes containing induced contact perturbations followed by corrective recovery ($326{,}972$ demonstration steps retained after role-masking out induced failure-setup prefixes). Physics-replay changepoint verification over contact force and joint displacement identifies $2{,}848$ probe and post-contact feedback boundaries and yields $327{,}637$ sliding $8$-step training windows, of which $5.13\%$ to $8.06\%$ in recovery trajectories straddle a physical feedback transition.

\begin{table}[t]
\centering
\caption{LIBERO-SelfTest Replay-Verified Corpus Statistics Across All 130 LIBERO Tasks.}
\label{tab:app-corpus}
\scriptsize
\setlength{\tabcolsep}{2.6pt}
\begin{tabular}{lcccccc}
\toprule
Suite & Tasks & Ep. & Pass / Rec. & Med.\ St. & Cross-$t_{\mathrm{fb}}$ & 2-Fing. \\
\midrule
Spatial & 10 & 80 & 40 / 40 & 298.5 & 6.64\% & 45.89\% \\
Object & 10 & 80 & 40 / 40 & 303.0 & 7.60\% & 53.42\% \\
Goal & 10 & 80 & 42 / 38 & 273.5 & 8.06\% & 35.00\% \\
Long-90 & 90 & 720 & 369 / 351 & 323.5 & 6.68\% & 44.33\% \\
Long-10 & 10 & 80 & 41 / 39 & 434.5 & 5.13\% & 38.35\% \\
\midrule
Total & 130 & 1{,}040 & 532 / 508 & -- & -- & -- \\
\bottomrule
\end{tabular}
\vspace{2pt}

\parbox{\linewidth}{\scriptsize\emph{Note:} Cross-$t_{\mathrm{fb}}$, fraction of $8$-step recovery windows straddling feedback transition $t_{\mathrm{fb}}(t)$; 2-Fing., antipodal two-finger contact fraction.}
\end{table}

\begin{table*}[t]
\centering
\caption{Extended Asynchronous, Non-Stationary Jitter, and Curved-Rim Contact Evaluation Across $0$--$400\,\mathrm{ms}$ Lag.}
\label{tab:app-async-300ms}
\scriptsize
\setlength{\tabcolsep}{3.3pt}\renewcommand{\arraystretch}{0.88}
\begin{tabular}{ll c cccccc cccccc}
\toprule
& & \multicolumn{1}{c}{$0\,\mathrm{ms}$} & \multicolumn{6}{c}{$200\,\mathrm{ms}$ ($d = 4$, $\gamma_{\mathrm{stab}} = 0.51$)} & \multicolumn{6}{c}{$300\,\mathrm{ms}$ ($d = 6$, $\gamma_{\mathrm{stab}} = 0.45$)} \\
\cmidrule(lr){3-3} \cmidrule(lr){4-9} \cmidrule(lr){10-15}
Model / Backbone & Suite (Role) & Sync. & Raw & Iss. & VLASH & Pref.-Al. & Fl.-Blend & Fl.+Pref. & Raw & Iss. & VLASH & Pref.-Al. & Fl.-Blend & Fl.+Pref. \\
\midrule
SmolVLA-450M & 4-suite mean & 75.8 & 25.8 & 28.3 & 50.0 & \textbf{78.3} & 59.2 & 75.8 & 11.7 & 10.0 & 30.0 & 74.2 & 58.3 & \textbf{75.8} \\
OpenVLA-OFT 7B & Spatial (SFT) & 94.0 & 32.0 & 68.0 & 60.0$^\dagger$ & 90.0 & \textbf{94.0} & 90.0 & 10.0 & 28.0 & 35.0$^\dagger$ & 70.0 & \textbf{92.0} & 72.0 \\
\quad Median $r_{\mathrm{contact}}$ & SFT ($\mathrm{mm}$) & 43.3 & 65.8 & 53.1 & -- & \textbf{43.4} & 48.2 & 44.5 & 82.4 & 56.7 & -- & 48.6 & \textbf{48.2} & 48.6 \\
OpenVLA-OFT 7B & Spatial (RIPT) & \textbf{96.0} & 26.0 & 64.0 & 70.0$^\dagger$ & 90.0 & \textbf{92.0} & 86.0 & 6.0 & 42.0 & 30.0$^\dagger$ & 72.0 & \textbf{90.0} & 74.0 \\
\quad Median $r_{\mathrm{contact}}$ & RIPT ($\mathrm{mm}$) & 43.5 & 68.5 & 54.6 & -- & \textbf{45.6} & 49.4 & 45.8 & 84.6 & 59.0 & -- & 50.8 & \textbf{49.4} & 50.3 \\
\midrule
\multicolumn{15}{l}{\emph{Non-stationary jitter on SmolVLA-450M (cols.\ 4--9 / 10--15: Raw, Fixed-$\bar d$ Pref., EAPN~\cite{eapn2026}, Train.-RTC~\cite{training_rtc2025}, Fl.+Pref., Fl.+Pref.+EAPN)}} \\
Step / Burst & $2{\to}6$ / $\{2,7\}$ & 75.8 & 38.3 & 56.7 & 60.0 & 65.0 & 71.7 & \textbf{75.8} & 42.5 & 60.8 & 55.0 & 54.2 & 70.8 & \textbf{76.7} \\
\bottomrule
\end{tabular}
\vspace{-4pt}
\end{table*}

\subsection{Policy Architectures and Post-Training Hyperparameters}
{\emergencystretch=1.6em
To isolate chunk-loss geometry from visual perception noise in Table~\ref{tab:takeover}A, the diagnostic \texttt{L1-ChunkNet} and \texttt{Flow-ChunkNet} MLPs ($4$ residual layers, $256$ hidden units, SiLU, layer normalization) receive low-dimensional simulator contact state, end-effector pose/velocity, and short-horizon force-displacement history, whereas all visuomotor policies (\texttt{TaskFiLM-ResNet18} in Table~\ref{tab:takeover}B and SmolVLA-450M in Table~\ref{tab:smolvla-bifurc}) consume only RGB views and proprioception with zero force inputs. Both policies train for $2{,}000$ AdamW steps (learning rate $3 \times 10^{-4}$, batch size $128$, three random seeds), with \texttt{Flow-ChunkNet} integrating $10$ midpoint Euler steps during inference. In Table~\ref{tab:takeover}B, \texttt{TaskFiLM-ResNet18} encodes dual $256 \times 256$ workspace and wrist RGB views with $64$-dimensional task-FiLM conditioning and $9$-dimensional proprioception into an $8 \times 7$ regression head (AdamW, learning rate $2 \times 10^{-4}$, batch size $64$). For pretrained SmolVLA-450M~\cite{smolvla2025} in Table~\ref{tab:smolvla-bifurc}, subspace weight interpolation~\cite{wiseft2022} blends fine-tuned weights separately across state/action projection heads and transformer blocks, while stalled prefix alignment gates index skipping on a realized in-flight displacement threshold of $4.5\,\mathrm{mm}$.

For the 7B VLA post-training audit in Table~\ref{tab:rl-retry}a, we evaluate \texttt{moojink/\allowbreak openvla-7b-\allowbreak oft-\allowbreak finetuned-\allowbreak libero-\allowbreak spatial} ($7.65\mathrm{B}$ parameters, $32$-layer Llama-2 backbone, $8 \times 7$ continuous action head with Laplace sampling scale $1.15$) before and after RIPT-VLA LoRA reinforcement post-training~\cite{ript2025} ($439$ rank-$32$ LoRA adapters totaling $110.8\mathrm{M}$ trainable parameters). Initial arm joint configurations receive paired uniform perturbations over $[-0.14, +0.14]\,\mathrm{rad}$ across five states per task over all ten LIBERO-Spatial tasks ($50$ states, eight stochastic rollouts per state, $260$-step horizon), and $3\mathrm{D}$ approach corridors are clustered via $15\,\mathrm{mm}$ RMS Euclidean distance across steps $24$ to $96$. Blockwise LoRA interpolation scales adapter weights by $0.50$ on early blocks ($0$--$11$) or mid-to-late blocks ($16$--$31$), whereas opening handoff disables RL LoRA adapters across the first $40$ control steps ($5$ chunks covering pre-contact corridor commitment on LIBERO-Spatial; replaceable on variable-length tasks by handing off at initial grasp-proximity deceleration or jaw-closure onset) before restoring full RL weights with exact MuJoCo state continuity.\par}

\section{Extended Empirical Evaluations}
\label{app:extended}

\subsection{Sawtooth Calibration, \texorpdfstring{$300$--$400\,\mathrm{ms}$}{300--400 ms} Lag, and Transport Jitter}
\label{app:ext-async}
\begin{figure}[!htbp]
\centering
\includegraphics[width=0.82\columnwidth]{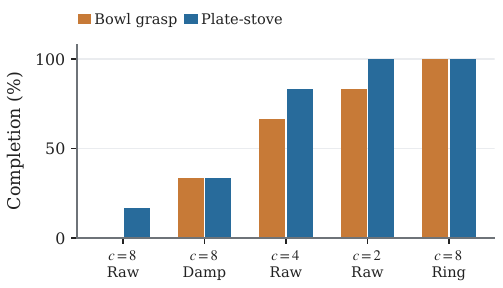}\vspace{-4pt}
\caption{Controlled MuJoCo sawtooth-delay completion ($8$-step chunk stride under two-step base lag) on bowl-grasp and plate-placement tasks comparing raw open-loop execution, proportional damping, stride reduction, and age-matched ring-buffer pose calibration.}
\label{fig:app-sawtooth}
\end{figure}

Controlled MuJoCo rollouts under periodic sawtooth observation age (two-step base lag with an $8$-step chunk stride; Fig.~\ref{fig:app-sawtooth}) isolate within-chunk lag growth prior to contact: raw $8$-step execution drops to $0.0\%$ completion on bowl grasping and $16.7\%$ on plate placement ($71.9$ and $80.7\,\mathrm{mm}$ terminal overshoot), whereas age-matched ring-buffer pose calibration restores $100.0\%$ completion on both tasks ($15.2$ and $12.2\,\mathrm{mm}$ overshoot).

Extending closed-loop VLA evaluation to $300$--$400\,\mathrm{ms}$ lag ($6$ to $8$ control steps) in Table~\ref{tab:app-async-300ms} further separates solver acceleration from external queue delay. Compressing flow-integration steps (FASTER~\cite{faster2026}) reduces model compute time without compensating transport lag ($30.8\%$ under $200\,\mathrm{ms}$ delay), while test-time prefix guidance (Guided RTC~\cite{rtc2025}) adds backpropagation overhead ($542.8$ vs.\ $353.2\,\mathrm{ms}$). On $7.65\mathrm{B}$ OpenVLA-OFT, uncompensated $300\,\mathrm{ms}$ lag pushes initial rim contact outside the $52.3\,\mathrm{mm}$ self-locking radius ($82.4$ and $84.6\,\mathrm{mm}$ median contact radius for SFT and RIPT), whereas Floquet blending pulls initial contact back inside the self-locking disk ($48.2$ and $49.4\,\mathrm{mm}$) and restores $92.0\%$ and $90.0\%$ completion. Under mid-episode delay step shifts and bimodal congestion bursts on SmolVLA-450M, indexing Floquet blending and prefix alignment by realized arrival age complements correlated flow noise (EAPN~\cite{eapn2026}), reaching $75.8\%$ and $76.7\%$ completion while reducing boundary splice slip to $1.7\%$ and $5.8\%$.

\subsection{Multi-Curvature Self-Locking, Wrist-Yaw Sensitivity, and Viscous-Spring Probing}
\label{app:ext-mechanics}
\label{app:ext-probing}
Table~\ref{tab:app-curved} validates the analytical quasi-static self-locking boundary (Eqs.~\eqref{eq:friction-cone} and~\eqref{eq:app-mechanics}) across $576$ controlled $500\,\mathrm{Hz}$ MuJoCo grasp trials spanning four smooth rim curvatures ($40$ to $95\,\mathrm{mm}$) and four friction coefficients ($0.30$ to $0.90$), matching empirical slip thresholds within $0.46\,\mathrm{mm}$ mean absolute error ($1.02\,\mathrm{mm}$ worst-case error). Sweeping wrist-yaw misalignment from $0^\circ$ to $-40^\circ$ confirms the asymmetric single-finger wrench mechanism in Eq.~\eqref{eq:app-mechanics}: early single-pad touch extends single-finger contact duration from $3.1$ to $7.3$ steps and drives grasp completion monotonically from $63.6\%$ down to $0.0\%$. On concave C-shaped handles where direct upward pulling snags the outer lip ($0/24$ poses cleared), routing end-effector motion along a subcell Shortley--Weller $2\mathrm{D}$ harmonic potential field~\cite{connolly1993harmonic,shortley1938numerical} clears all $24/24$ snag configurations without collision.

\begin{table}[!htbp]
\centering
\caption{Multi-Curvature MuJoCo Self-Locking Critical Offset (mm) Across 576 Trials.}
\label{tab:app-curved}
\scriptsize
\setlength{\tabcolsep}{2.5pt}
\begin{tabular}{lcccc}
\toprule
Radius Profile & $\mu = 0.30$ & $\mu = 0.50$ & $\mu = 0.70$ & $\mu = 0.90$ \\
\midrule
$40\,\mathrm{mm}$ (neck) & 11.24 / 10.73 & 16.52 / 17.54 & 22.68 / 22.95 & 26.20 / 27.06 \\
$55\,\mathrm{mm}$ (mug) & 14.93 / 14.56 & 23.55 / 23.80 & 30.94 / 31.14 & 35.86 / 36.72 \\
$75\,\mathrm{mm}$ (bowl) & 19.85 / 19.68 & 31.75 / 32.15 & 41.95 / 42.07 & 48.75 / 49.60 \\
$95\,\mathrm{mm}$ (wide) & 24.77 / 24.79 & 39.95 / 40.51 & 52.97 / 53.00 & 61.63 / 62.49 \\
\bottomrule
\end{tabular}
\vspace{2pt}

\parbox{\linewidth}{\scriptsize\emph{Note:} Empirical / analytical quasi-static critical offset $h_{\mathrm{crit}}$ ($0.46\,\mathrm{mm}$ MAE, $1.02\,\mathrm{mm}$ max error across $16$ curvature--friction pairs).}
\end{table}

Across $7{,}200$ compound articulated-joint rollouts combining $200$--$300\,\mathrm{ms}$ observation lag with unseen Coulomb and non-Coulomb viscous-spring resistance, pairing Floquet-damped approach control with post-deadband sequential probing and closure target withdrawal sustains $79.5\%$ completion under $200\,\mathrm{ms}$ Coulomb friction and $77.0\%$ under viscous-spring resistance while cutting safety overloads by more than half.

\subsection{Exhaustively Enumerable Symbolic Planning Audit on \emph{Countdown}}
\label{app:countdown}
\begin{figure}[!t]
\centering
\includegraphics[width=0.78\columnwidth]{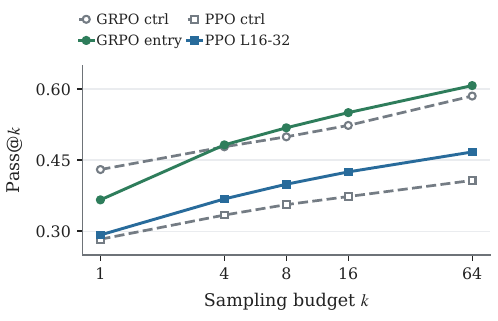}\vspace{-4pt}
\caption{Enumerable \emph{Countdown} Pass@$k$ scaling ($135$ held-out tasks) confirming opening-branch localization under full-parameter PPO and GRPO.}
\label{fig:app-countdown}
\end{figure}

Because continuous $3\mathrm{D}$ manipulation workspaces do not permit exhaustive enumeration of all valid unvisited trajectories, we complement our 7B VLA post-training audit with $135$ train-disjoint \emph{Countdown} arithmetic planning problems~\cite{pan2025countdown} on $36$-layer Qwen2.5-3B ($64$ rollouts per problem sampled with temperature $0.7$ and top-$p$ $0.9$). Canonicalizing commutative operations partitions the exact solution tree of each problem into discrete entrance families defined by the opening pair of operands and arithmetic operator, enabling exact measurement of canonical solution coverage alongside unbiased $\mathrm{Pass@}1$ and $\mathrm{Pass@}64$. As shown in Table~\ref{tab:app-countdown} and Fig.~\ref{fig:app-countdown}, full-parameter PPO and GRPO improve $\mathrm{Pass@}1$ while contracting $64$-sample canonical branch coverage to $0.118$ and $0.203$. Attenuating RL updates across mid-to-late transformer blocks (layers $16$--$32$) or gating interpolation strictly across the opening arithmetic step recovers canonical branch coverage ($0.203 \to 0.289$) and $\mathrm{Pass@}64$ ($0.585 \to 0.607$) without retraining, whereas initializing GRPO from a multi-solution SFT prior ($4$ distinct valid traces per training problem) preserves broad branch coverage ($0.603 \to 0.595$) alongside higher accuracy.

\begin{table}[!htbp]
\centering
\caption{Enumerable Symbolic Planning Audit (Countdown, $135$ Held-Out Problems, $k=64$).}
\label{tab:app-countdown}
\scriptsize
\setlength{\tabcolsep}{2.4pt}
\begin{tabular}{lccc}
\toprule
Condition ($T = 0.7$, top-$p = 0.9$) & Pass@1 & Pass@64 & Sol.\ cov.@64 \\
\midrule
PPO step 275 (ctrl) & 0.283 & 0.407 & 0.118 \\
PPO L0--12 ($\alpha = 0.50$) & 0.282 & 0.415 & 0.124 \\
PPO L16--32 ($\alpha = 0.50$) & \textbf{0.292} & \textbf{0.467} & \textbf{0.177} \\
\cmidrule(lr){1-4}
GRPO step 450 (ctrl) & \textbf{0.430} & 0.585 & 0.203 \\
GRPO stat.\ / entry L16--32 ($\alpha = 0.70$) & 0.316 / 0.366 & \textbf{0.630} / 0.607 & \textbf{0.340} / 0.289 \\
\cmidrule(lr){1-4}
SFT ($M = 4$) / + GRPO & 0.307 / \textbf{0.321} & 0.869 / \textbf{0.874} & \textbf{0.603} / 0.595 \\
\bottomrule
\end{tabular}
\end{table}

\end{document}